# FULL HIGH-DIMENSIONAL INTELLIGIBLE LEARNING IN 2-D LOSSLESS VISUALIZATION SPACE

Boris Kovalerchuk, Hoang Phan

Dept. of Computer Science, Central Washington University, USA
Boris.Kovalerchuk@cwu.edu, Hoang.Phan@cwu.edu

**Abstract**. This study explores a new methodology for machine learning classification tasks in 2-dimensional visualization space (2-D ML) using Visual knowledge Discovery in lossless General Line Coordinates. It is shown that this is a full machine learning approach that does not require processing n-dimensional data in an abstract n-dimensional space. It enables discovering n-D patterns in 2-D space without loss of n-D information using graph representations of n-D data in 2-D. Specifically, this study shows that it can be done with static and dynamic In-line Based Coordinates in different modifications, which are a category of General Line Coordinates. Based on these inline coordinates, classification and regression methods were developed. The viability of the strategy was shown by two case studies based on benchmark datasets (Wisconsin Breast Cancer and Page Block Classification datasets). The characteristics of page block classification data led to the development of an algorithm for imbalanced high-resolution data with multiple classes, which exploits the decision trees as a model design facilitator producing a model, which is more general than a decision tree. This work accelerates the ongoing consolidation of an emerging field of full 2-D machine learning and its methodology. Within this methodology the end users can discover models and justify them as self-service. Providing interpretable ML models is another benefit of this approach.

**Keywords**. Interpretable machine learning, classification, regression, visual knowledge discovery.

## 1. Introduction

Today, the Machine Learning (ML) field is heavily focused on **Interpretable Machine Learning (IML)** [14]. While the term IML is popular several other terms like explainable, intelligible, comprehensible and others often are used interchangeably or with some nuanced differences. These methods cover everything from constructing explainable models from scratch to explaining black box models. Using visual methods to develop explainable machine learning models is one of the appealing possibilities. It is difficult, though, because machine learning data are multidimensional and which we cannot see by the naked eye. Therefore, new methods to make this possible are required.

Traditional techniques are lossy, not retaining all multidimensional information while converting n-D data to two dimensions. **General Line Coordinates (GLC)** representation of n-D data, in contrast, is lossless [12]. With no information loss, this visual representation made it possible to perform **full multidimensional machine learning in two dimensions**. This opens the opportunity for the end user to be involved in the entire machine learning process because of its visual nature in 2D. It dramatically drops the requirement of deep mathematical knowledge of ML processed from the end users. There are two benefits to this strategy. In straightforward circumstances, a user can find the pattern visually by simply looking at the data represented in GLC. A user can find patterns in 2-D representations by applying new 2-D ML techniques in more complicated scenarios. This area is developing now into an entirely new branch of machine learning.

This chapter falls within this category and concentrates on new **In-line Coordinates (ILC)** expanding [7]. These coordinates are a subset of General Line Coordinates [12] and are defined in section 2. The viability of full 2-D ML with various General Line Coordinates (Shifted Paired Coordinates, Elliptic Paired Coordinates, CPC-R and GLC-L) was demonstrated in a number of earlier works [5, 7, 11, 13, 15, 21]. When it comes to 2-D modeling of non-image data, this field of study may be traced back to [3] for Parallel Coordinates and [18]. Often, 2-D studies in ML only use straightforward 2-dimensional examples to graphically demonstrate ML algorithms. The exploration of Parallel Coordinates in visual analytics studies has been quite active for clustering-related tasks [17], but much less so for supervised learning, which is the subject of our study. Among the relevant works are [1, 19, 22].

This research expands [7] with new case studies and algorithms for more challenging data, which include the heavily imbalanced high-resolution data with multiple classes like Page Block Classification data. The algorithm uses a decision tree as a model design facilitator for producing a model, which is more general than a decision tree.

We propose a **full 2D ML methodology** expanding [7] to consolidate prior investigations based on General Line Coordinates into a single, overarching paradigm. In the past, 2D machine learning research was viewed as merely **auxiliary** exploratory data/model visualization occurring primarily after or before the actual machine learning due to loss of n-D information. It was assumed that since loses of n-D information in 2-D, comprehensive interpretable n-dimensional analysis in n-D space is required to build ML models. The full 2-D ML approach demonstrates that it is not required. The **main contribution** is the methodology that goes *beyond a range of methods of lossless visualization* methods developed in [12] to multiple *machine learning techniques for discovering complete n-D patterns in 2-D visualization space* broadening human-assisted visual pattern discovery. Relative to the common lossy visualization of n-D data, the contribution is in abilities to actual *discovering complete n-D patterns in 2-D lossless visualization space.*

This chapter is structured as follows. In-line coordinates' primary notions are defined in Section 2. The Box Classification (BC), Linear Classification, and Regression algorithms are presented in Section 3. The case study utilizing benchmark Wisconsin Breast Cancer (WBC) data that shows the viability of the strategy is covered in Section 4. Section 5 presents the results with Page Block Classification (PBC) data and BC algorithm for imbalanced high-resolution data with multiple classes and compares results with other algorithms. The conclusions are presented in Section 6.

## 2. In-line based coordinate systems

With the General Line Coordinates defined in [12], $n$ coordinate axes can be drawn in 2-D in a variety of ways, including curved, parallel, unparalleled, collocated, unconnected, etc. Testing several GLC coordinate mappings and orders are needed to discover the best visual representation of n-D data possible.

GLCs contain In-line Coordinates, which are similar to Parallel Coordinates but differ in that their axes are horizontal rather than vertical (see Fig. 1). On the same line, all coordinates $X_1, X_2, \ldots X_n$ are collocated and may or may not overlap. The condition of lossless representation of n-D point $\mathbf{x}=(x_1, x_2, \ldots, x_n)$ in 2-D is satisfied by a sequence of directed curves or polylines, which connect adjacent points $x_i$, $x_{i+1}$ in coordinates $X_i$, $X_{i+1}$ as shown in Fig. 1. These curves/polylines of various heights and shapes form graphs [12]. Both Parallel Coordinates and In-line Coordinates require the same number of nodes and links and can be used similarly.

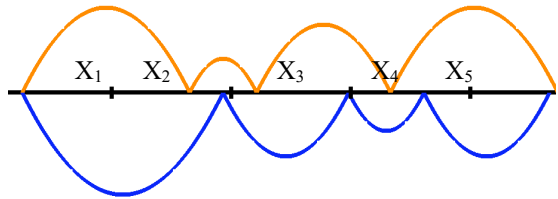

Fig. 1. Two 5-D points of two classes in In-line Coordinates.

The connections between the nodes are directed edges, however when the direction matches the order of coordinates, arrowheads are not necessary. It is possible to draw n-D points from distinct classes above and below the coordinate line to more clearly illustrate the differences between them (see Fig. 1). Three ILC location modes are considered:

(L1) *Sequential* ILC with coordinates located one after another (Fig. 1).

(L2) *Collocated* ILC with coordinates drawn at the same location with full overlap (Fig. 2a).

(L3) *Generic* ILC where some coordinates can be *sequential*, *collocated*, *overlapping,* or *disjoined* (Fig. 2b).

(L4) Dynamic ILC with coordinated located *dynamically* as explained later.

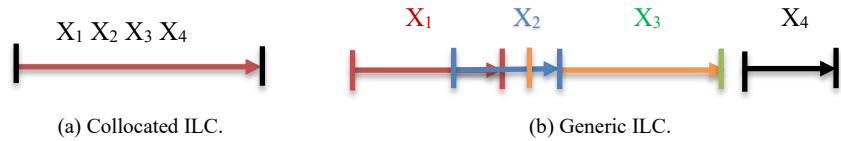

(a) Collocated ILC.          (b) Generic ILC.
Fig. 2. Options to locate coordinates in In-line Coordinates.

By choosing a certain ILC overlap in L3, a given n-D point **c** can be collapsed to a single 2-D point while remaining fully recoverable. When n-D point **c** is the class's center point and other n-D points are concentrated around it, this is a useful visual characteristic. Another method for enhancing the visibility of the patterns of interest is to reorder the coordinates $X_1$-$X_n$.

There are numerous ways to build links connecting points $x_i$ on coordinates $X_i$ by giving meaning to their width, height and other charateristics, to communicate more information. See an example in Fig. 3a for a 7-D point with values of $x_3$ and $x_4$ encoded as the **height** and the **width** of the line that connects $(x_1,x_2)$, and values of $x_6$ and $x_7$ the height and width of the line that connects $(x_2,x_5)$. This makes the ILC base line shorter because only the three coordinates $x_1$, $x_2$ and $x_5$ are directly encoded in it.

Fig. 3b and Fig. 4 show other options. Fig. 3b uses the lengths of **sides** of the line that connects points $x_1$ and $x_2$ to encode values of $x_3$ and $x_4$, instead of using its width and height. Similarly, lengths of sides of the line that connects points $x_2$ and $x_5$ encode values of $x_6$ and $x_7$. In-line Coordinates' primary objective is to facilitate the discovery of n-D patterns and rules with the highest levels of recall and precision.

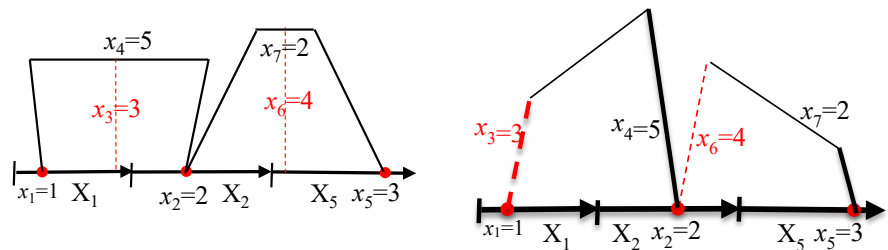

(a) $x_3,x_4,x_6$ and $x_7$ encoded by the height and width of the link lines that connect $(x_1,x_2)$ and $(x_2,x_5)$.    (b) $x_3,x_4,x_6$ and $x_7$ encoded by of length of sides of the link lines that connect $(x_1,x_2)$ and $(x_2,x_5)$.
Fig. 3. 7-D point **x** = $(x_1,x_2,x_3,x_4,x_5,x_6,x_7)$ = (1,2,3,5,3,4,2) in two ILBCs.

**In-Line Based Coordinates (ILBC)** is the name given to these visual representations since they no longer use a single baseline as a basic ILC representation. Instead, they are based on it.

Fig. 4a simplifies Fig. 3b, by making sides **vertical,** and Fig. 4b simplifies this Fig, 4a further by removing vertical lines, which go down to the baseline and keeping only solid lines. In this figure we use more generic simplified notation with attributes named from *a* to *g*, because any of the coordinates $\{x_i\}$ can be assigned to be on the baseline or on link lines and in any order.

In contrast to Parallel Coordinates, which need seven nodes and six edges, this figure enables the complete restoration of all seven values and only needs four nodes and three edges. The vertical sides of the image in Fig. 4 can be understood as follows. The ILC baseline occupies the Cartesian x-coordinate, while all vertical values *a* through *g* are situated on their corresponding coordinates *A* through *G*, which are vertically collocated on. A **combination of two ILCs—horizontal and vertical—** makes up ILBC in Fig. 4. Such coordinates will be denoted by **ILC2.**

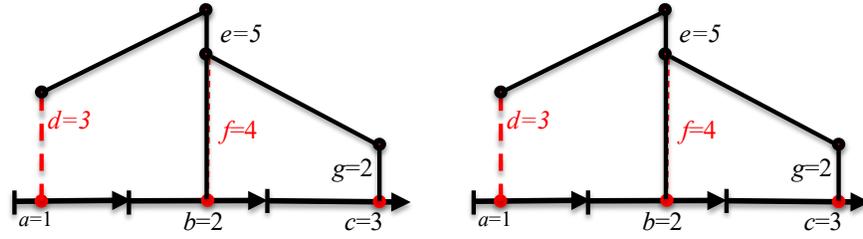

(a) Vertical simplification of Fig. 3b.   (b) Minimized representation.
Fig. 4. 7-D point **x** = (*a,b,c,d,e,f,g*) = (1,2,3,5,3,4,2) in ILC2 with vertical sides.

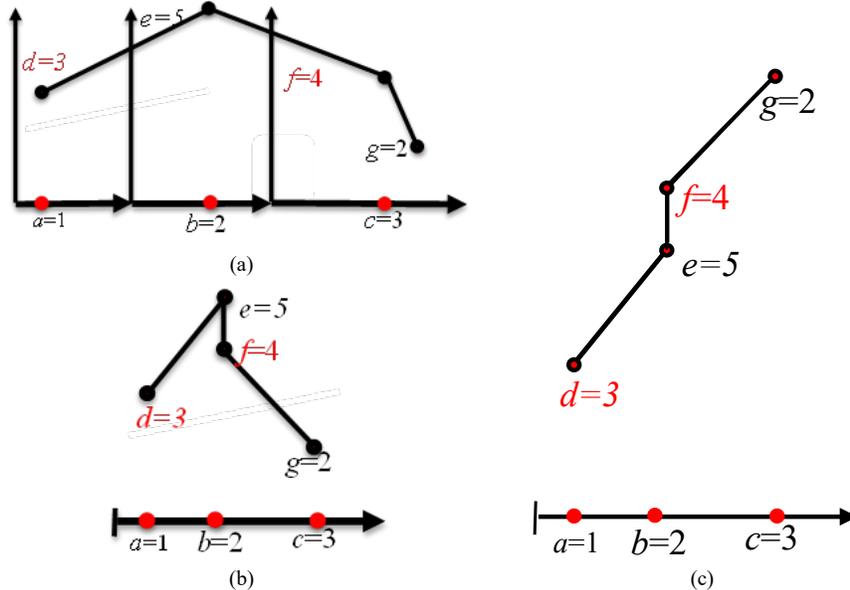

Fig. 5. (a) 7-D point **x** = (*a,b,c,d,e,f,g*) = (1,2,3,5,3,4,2) in SPC, (b) in ILBC partial dynamic, (c) in fully dynamic.

Next, ILC2 is compared with Shifted Paired Coordinates (SPC) [12] on the same 7-D point shown in Fig. 5. In Fig. 5a SPC *d*, *e*, *f*, and *g* are also vertical, but start at the origins of individual horizontal coordinates, which are paired. SPC also requires 4 nodes and 3 edges that are longer than in ILBC in Fig. 4b. In ILC and ILBC above, the location of all coordinates on the horizontal baseline is fixed with their values located on this baseline. It is called a **static mapping** [12]. In contrast in the **dynamic mapping** of the given n-D point **x**, the location of the next value $x_{i+1}$ in its 2-D graph **x*** *depends on the location and value of prior $x_i$*. It is a common concept for all dynamic General Line Coordinates [12], not only ILC and ILBC.

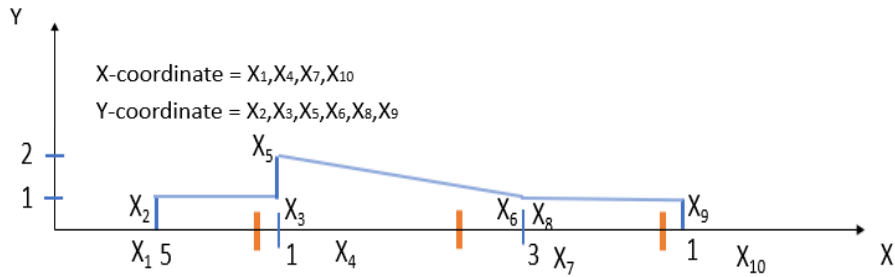

(a)

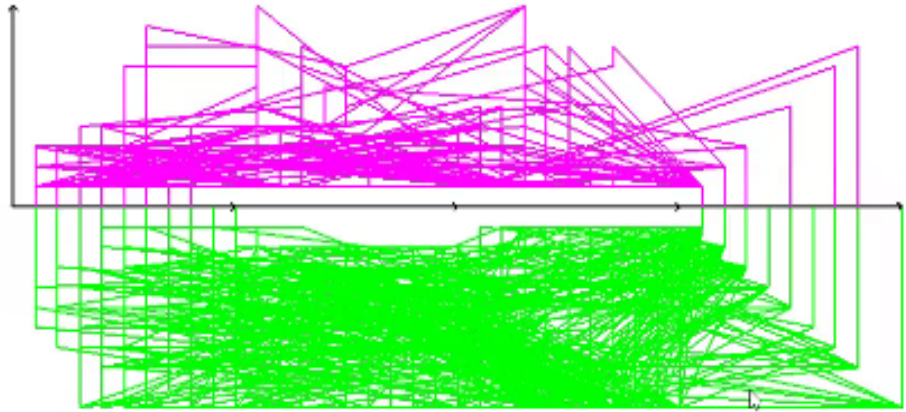

(b)
Fig. 6. WBC data in partial dynamic ILBC, (a) a single case , (b) many cases of two classes "mirrored".

The same 7-D point is shown in Fig. 5b. Here the coordinate B starts at point $a=2$ of coordinate A with value $b=2$ located at the distance 2 from point $a=1$. Respectively coordinate C starts at point $b=2$ with point $c=3$ located at the distance 3 from $b=2$. The respective vertical coordinates $d$, $e$, $f$, and $g$ start at the origin of the horizontal baseline. Thus, all of them are **collocated and static**. Here horizontal coordinates are dynamic, but vertical coordinates are static, therefore, ILBC is called **partial dynamic ILC2**. Fig. 5c shows a **full dynamic ILC2** where vertical coordinates are dynamic in the same ways as horizontal coordinates, where the location of $e$, $f$ and $g$ points depends on the location of their prior points.

Figs. 6 and 7 show Wisconsin Breast Cancer (WBC) data of two classes in ILC, where vertical coordinates are collocated and horizontal are static. Figs. 6a and 7a are example of one case from WBC dataset for Figs. 6b and 7b. In Fig. 6a, the X-coordinate represents the values of $x_1$, $x_4$, $x_7$, and $x_{10}$. The y-coordinate represents the values of $x_2$, $x_3$, $x_5$, $x_6$, $x_8$, and $x_9$. These values are $x_1$=5, $x_2$=1, $x_3$=1, $x_4$=1, $x_5$=2, $x_6$=1, $x_7$=3, $x_8$=1, $x_9$=1, and $x_{10}$=1. All patients in WBC dataset are represented in this manner.

Drawing classes "mirrored" in Fig. 6 allows to compare and see the difference and similarities of patterns of two classes without their occlusion. Fig. 7b shows much better separation of WBC classes in fully dynamic ILC2.

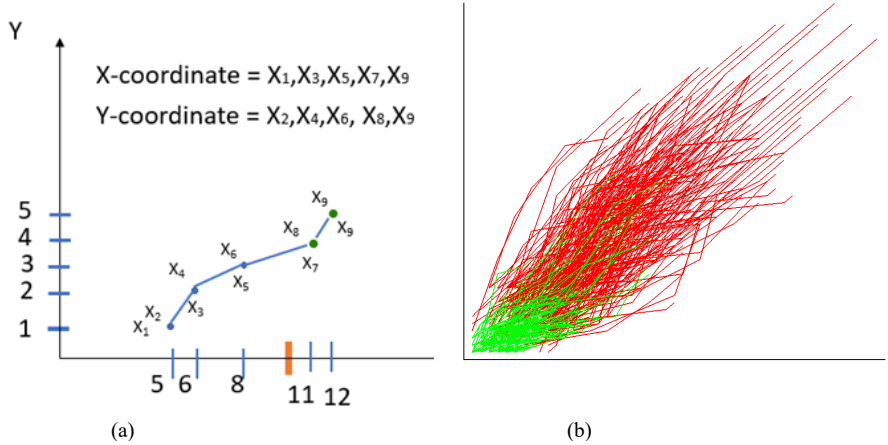

(a)                          (b)
Fig. 7. (a) Example coordinates. (b) Wisconsin Breast Cancer dataset in fully dynamic ILC2.

The general formula to locate pairs $(x_i, x_j)$ in **partial dynamic IL2** is given by mapping $L$ as follows:

$L(x_i, x_j) = (x_1+x_3+…+ x_i, x_j)$.

Respectively the general formula to locate pairs $(x_i, x_j)$ in **fully dynamic ILBC** is given by mapping $L$ as follows:

$L(x_i, x_j) = (x_1+x_3+…+ x_i, x_2+x_4+…+ x_j)$.

Example: In Fig. 7a, the value (5,1) represents $x_1$ and $x_2$. The value (6,2) presents $x_{3s}$ and $x_{4s}$ because the position of $x_{3s}$ depends on the location of $x_1$. This means that to retrieve the true value of $x_3$, the value of $x_1$ is subtracted from $x_{3s}$. Therefore, $x_3 = x_{3s} - x_1 = 6 - 5 = 1$. This process is repeated for all remaining points. With that, the drawn $x$ values are (5,1,6,2,8,3,11,4,12,5), and the true $x$ values are (5,1,1,1,2,1,3,1,1,1).

Next, **weighted dynamic ILBC** is introduced which is given by mapping $L_w$:

$$L_w(x_i, x_j) = (w_1x_1+ w_3x_3+\ldots+ w_ix_i, w_2x_2+ w_4x_4+\ldots+ w_jx_j),$$

where W={$w_i$} is a set of weights assigned to coordinates.

## 3. Classification and regression algorithms with in-line coordinates

## 3.1. Box Classification algorithm

The main idea of the **Box Classification (BC) algorithm** is finding a good box with high purity and many cases, then remove all cases, which are in this box, and repeat the process of finding other good boxes in remaining cases and continue this process until all cases of all classes will be in one of the good boxes. This process is interactive and partially automated. Automation includes computing parameters of the candidates for the good boxes.

The BC algorithm operates on n-D data visualized in ILBC in the following main steps.

Step 1: **Split data** to training and validation data.

Step 2: **Search/discover "good" boxes** $B_i$ and form rules on training data in ILBC visualizations that cover dataset cases as pure as possible. Good box' criterion is usually decided by number of cases that it covers and its purity.

Step 3: **Test BC** rules on independent data.

Step 4: **Pruning** a set of discovered rules to decrease overfitting.

Below we discuss details of these steps. For many datasets, the *exhaustive grid search* is possible in Step 2. For instance, it was used to discover boxes with WBC dataset. With WBC dataset, a "good" box is discovered when it covers at least more than 10% of cases of the remaining data. The search of boxes and rules is a *sequential hierarchical process for each class.* If there are multiple "good" boxes, then these "good" boxes are ranked based on the purity of the boxes. **Purity** of the box is the ratio of the number of cases of the *dominant class* to the number of cases of all other classes in the box. The *presence* of the case in the box is identified by the fact that a *node* of the graph of the n-D point **x** is in the box B. In the rule below, it will be denoted **x**∈ B for short. This definition leads to simpler interpretation of rules based on such boxes, than an alternative definition, which requires that just an *edge* to graph **x*** of the n-D point **x** crosses the box $B_i$.

To form BC rules with the discovered boxes the algorithm starts with discovering basic rules:

$$R_i: \text{if } \mathbf{x}\in B_i \Rightarrow \mathbf{x}\in \text{Class } C_i. \tag{1}$$

and then discovers more complex BC rules sequentially

$$R_i: \text{if } \mathbf{x}\in B_i \ \& \ \mathbf{x}\notin (B_m \cup B_p \cup\ldots\cup B_t) \Rightarrow \mathbf{x}\in \text{Class } C_i \tag{2}$$

Here $B_i$ is a current "good" box, and other boxes are prior "good" boxes with cases from these boxes removed before $B_i$ is searched. For WBC dataset, the algorithm found 13 rules as presented in Table 2 in Section 4.

The BC algorithm as an **iterative visual logical classifier** produces a series of rectangular areas (boxes). Fig. 8 shows multiple WBC cases visualized in IVLC with boxes found by the BC algorithm.

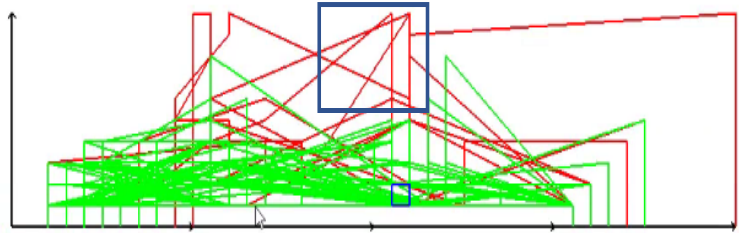

Fig. 8. Examples of boxes discovered in Wisconsin Breast Cancer dataset.

The main steps of discovering boxes in **Step 2** are:

Step 2.1. Create a *grid* in the ILC area. Each cell of the *grid* is a basic *box*. The box size is increased vertically and then horizontally.

Step 2.2. If the number of boxes (grid cells) and the number of n-D points is relatively small compute purity of *each* box. For the large number of boxes and n-D points use optimization and heuristic algorithms such as genetics algorithms.

Step 2.3. For each class create a list of boxes, where this class dominates. All boxes in this list must cover more cases than the desired percentage of the remaining cases. For instance, for WBC dataset it was 10% of the remaining data cases.

Step 2.4. For each class find a box of highest purity to classify new cases in this box.

Step 2.5. Create a classification rule with highest ranking box like in (1).

These steps in a pseudocode are below.

```
// Set up grid in ILC area
CreateGrid(size)

// Calculate purity of each box
For each box in grid
    CalculatePurity()

// Create list of boxes for each class
For each class
     CreateBoxList(class, coverage_threshold)

// Classify cases
For each class
    For each box in box list
        Use box with highest purity to classify cases
        UpdateClassificationRule(box, class)

// Return classification rule
Return ClassificationRule()
```

Example: for WBC dataset, there are 382 green cases and no red cases that cross box $B_1$ as shown in Fig. 10 in Section 4. Therefore, in this example, rule $R_1$ is created as follow,

$$R_1: \mathbf{x} \in B_1 \Rightarrow \mathbf{x} \in G \text{ (Benign, 382 cases).} \tag{3}$$

More boxes and rules for WBC data are identified later in Table 1 in Section 4.
Step 2.6. Exclude all cases that are in boxes used in steps 2.4 and 2.5.
Step 2.7. Conduct steps 2.2 - 2.6 for all remaining cases.

The **pruning step** is to deal with many "mini" boxes that contain less than a certain percentage of the cases of the original dataset with low level of generalization to avoid overfitting and data memorization. With WBC dataset, it was decided that if a box classified less than 17 cases of the total number of cases, which is 683, then it is very likely that this box is overfitting the data. Example: box $B_5$ classified only 14 red cases (about 2.05% of 683 cases). So, box $B_5$ is likely overfitted WBC data. This problem is also well known for decision trees. Without control the depth of decision tree and the number of cases in each terminal node, decision trees produce many terminal nodes with only few cases in each of them. The pruning of decision trees removes overfitting but decreases the accuracy of classification.

The step of the BC algorithm employs a **pruning** approach which
(a) associates "mini" boxes with the larger boxes interactively or
(b) refuses to predict cases that belong to "mini" boxes.

The association (a) is conducted as follows. Consider two boxes $B_1$ and $B_2$ for class $C_1$ the visualization allows to see their mutual location and to create a joint rule based on them. If boxes are adjacent a single bigger box $B_{1,2}$ is produced from them. If the boxes are not adjacent than a new rule is formed:

$$\text{If } \mathbf{x} \in B_1 \text{ or } \mathbf{x} \in B_2 \text{ then } \mathbf{x} \in C_1$$

The general form for of a new rule is,

$$\text{If } \mathbf{x} \in B_1 \cup B_2 \text{ then } \mathbf{x} \in C_1$$

The interactivity of the BC algorithm has an advantage of allowing the end-users to observe "mini" boxes and decide to follow (a) or (b).

### 3.2. Linear Classification and Regression algorithm

Design of rules based on boxes have limitations. One of them is **locality** of each box, i.e., the box covers only nearby cases. Typically, several boxes are needed to cover all data. In contrast, a single linear classifier can cover all data if data are linearly separable. The goal of this section is proposing an analog of a linear classifier in ILC. Fig. 9 illustrates the proposed approach. First, a black line is built which is used to project all cases of both classes to this line. If the projected endpoints of cases of one class C mostly concentrate on the one side of the discrimination blue line, then a linear discrimination model M is discovered, where $T$ is threshold on the black line shown with the blue discrimination line:

$$M(\mathbf{x}) > T \Rightarrow \mathbf{x} \in C \qquad (4)$$

Another more common linear discrimination model for two classes $C$ and $Q$ is:

$$M(\mathbf{x}) > T \Rightarrow \mathbf{x} \in C \text{ else } \mathbf{x} \in Q \qquad (5)$$

The model (4) only covers n-D points $\mathbf{x}$, where $M(\mathbf{x}) > T$. This is the situation in Fig. 9 for red class above the blue line. This means that if a single model (5) cannot be built, several models like (4) need to be built that may require different black lines, where endpoints are projected as shown in Fig. 9b. Moreover, it can relax a requirement that only endpoints are projected. It can project some intermediate nodes $x_k$ and $x_u$ of graph of $\mathbf{x}^*$ for $k < n$ and $u < n$, where $n$ is the dimension of n-D point $\mathbf{x}$ getting models like,

$$M(\mathbf{x}_k) > T \Rightarrow \mathbf{x} \in C.$$

$$M_1(\mathbf{x}_k) > T_k \ \& \ M_2(\mathbf{x}_u) > T_u \Rightarrow \mathbf{x} \in C.$$

Those intermediate points can be found by the BC algorithm presented above. Figs. 9a and 9b look different because the black lines are drawn differently, which allow to optimize the prediction parameters and accuracy. The endpoints of the black lines are chosen so that a higher precision accuracy can be obtained.

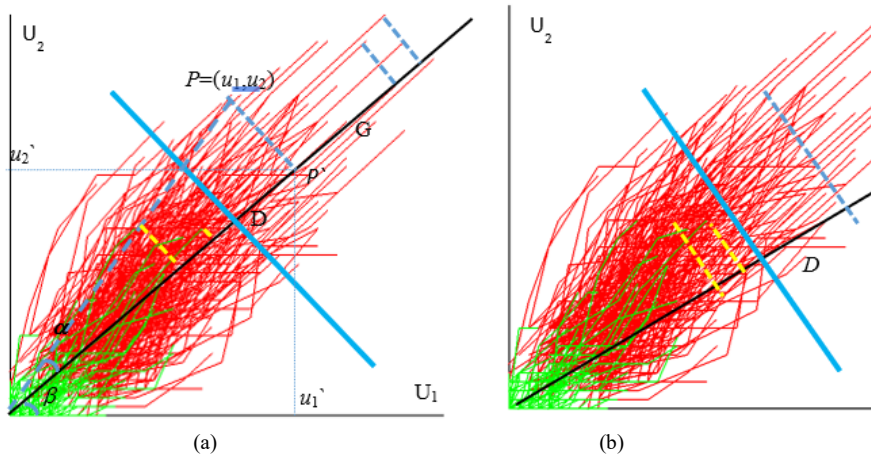

(a) (b)
Fig. 9. Linear Classification and Regression algorithm with projection line at different angles in full dynamic ILP2.

# 4. Case study with WBC data in ILC2

## 4.1. Experiment on all WBC data

This section presents the results of the computational experiment for discovering classification rules for WBC data encoded in partial dynamic ILC2 using the BC algorithm. The discovered 13 pure boxes are presented in Table 1. These rules can be simplified and pruned interactively. With WBC dataset, exhaustive grid search is used to discover the hyper-parameters (corners of the boxes), See column 2 and 5 in Table 1. Exhaustive grid search can be used with WBC dataset because WBC dataset has a low-resolution attributes with only ten values in each attribute.

With higher dimension and higher resolution dataset, exhaustive grid search will take much greater computational time because there will be much more cells in the grid. We avoided such computations by proposing a heuristic algorithm that is described in section V. The boxes cover all WBC cases. Hyper-parameters are values $x_1$, $x_2$, $y_1$, $y_2$, which identify left, right, bottom, and top corners of the box within ILC2 display.

Table 1. Discovered boxes ILC2.

| Box | $x_1, x_2, y_1, y_2$ | Cases | Box | $x_1, x_2, y_1, y_2$ | Cases |
|---|---|---|---|---|---|
| $B_1$ | 15,20.5,1,1.5 | 382 | $B_2$ | 23.5,39.5,8,5,10 | 166 |
| $B_3$ | 1,3.5,0.5,2 | 28 | $B_4$ | 20,22.5,6,6.5 | 26 |
| $B_5$ | 9.5,10,5,6.5 | 14 | $B_6$ | 16,21,0.5,2 | 18 |
| $B_7$ | 17.5,18.5,3,3.5 | 23 | $B_8$ | 14.5,17,2.5,3 | 7 |
| $B_9$ | 28.5,29,2.5,3.5 | 4 | $B_{10}$ | 17.5,18.5,3,3.5 | 10 |
| $B_{11}$ | 14.5,15,5.5,6 | 4 | $B_{12}$ | 26.5,27,7,7.5 | 1 |
| $B_{13}$ | 28,28.5,0.5,9.5 | 10 | | | |

Table 2 presents the rules constructed from these boxes in the hierarchical process of the BC algorithm that was described above. The benign class is denoted as B and is drawn as green with letter G used to identify this class in Table 2. Respectively, the class malignant is denoted by M and R (red) for short.

Table 2. Rules $R_1$-$R_{13}$ with precision P=100%.

| *Benign, B (green, G) class rules.* |
|---|
| $R_1$: $\mathbf{x} \in B_1 \Rightarrow \mathbf{x} \in G$ (382 cases) |
| $R_3$: $\mathbf{x} \in B_3 \Rightarrow \mathbf{x} \in G$ (28 cases) |
| $R_6$: $\mathbf{x} \in B_6$ & $\mathbf{x} \notin B_2 \cup B_4 \cup B_5 \Rightarrow \mathbf{x} \in G$ (18 cases) |
| $R_8$: $\mathbf{x} \in B_8$ & $\mathbf{x} \notin B_2 \cup B_4 \cup B_5 \cup B_7 \cup B_{10} \cup B_{13} \Rightarrow \mathbf{x} \in G$ (7 cases) |
| $R_9$: $\mathbf{x} \in B_9$ & $\mathbf{x} \notin B_2 \cup B_4 \cup B_5 \cup B_7 \cup B_{10} \cup B_{13} \Rightarrow \mathbf{x} \in G$ (4 cases) |
| $R_{11}$: $\mathbf{x} \in B_{11}$ & $\mathbf{x} \notin B_2 \cup B_4 \cup B_5 \Rightarrow \mathbf{x} \in G$ (4 cases) |
| $R_{12}$: $\mathbf{x} \in B_{12}$ & $\mathbf{x} \notin B_2 \cup B_4 \cup B_5 \cup B_7 \cup B_{10} \Rightarrow \mathbf{x} \in G$ (1 case) |
| *Malignant, M (red, R) class rules.* |
| $R_2$: $\mathbf{x} \in B_2 \Rightarrow \mathbf{x} \in R$ (166 cases) |
| $R_4$: $\mathbf{x} \in B_4 \Rightarrow \mathbf{x} \in R$ (26 cases) |
| $R_5$: $\mathbf{x} \in B_5$ & $\mathbf{x} \notin B_1 \cup B_3 \Rightarrow \mathbf{x} \in R$ (14 cases) |
| $R_7$: $\mathbf{x} \in B_7$ & $\mathbf{x} \notin B_1 \cup B_3 \cup B_6 \Rightarrow \mathbf{x} \in R$ (13 cases) |
| $R_{10}$: $\mathbf{x} \in B_{10}$ & $\mathbf{x} \notin B_3 \cup B_6 \cup B_8 \cup B_9 \Rightarrow \mathbf{x} \in R$ (10 cases) |
| $R_{13}$: $\mathbf{x} \in B_{13}$ & $\mathbf{x} \notin B_1 \cup B_3 \cup B_6 \cup B_8 \cup B_9 \cup B_{11} \cup B_{12} \Rightarrow \mathbf{x} \in R$ (10 cases) |

The boxes and rules in Tables 1 and 2 cover all 444 B cases and 239 M cases. Boxes $B_1$-$B_4$ and respective rules $R_1$-$R_4$) cover most of the cases (602 cases) with 100% precision without any misclassified cases. This means that 88.14% of all cases are classified by *simplest single box rules* without any other boxes involved. The other rules involve "negated" boxes requiring that the case does not belong to these boxes to satisfy the rule. Some rules in Table 2 such as rule $R_8$ have simplified forms too with reduced "negated" boxes, because all cases of some boxes are covered by other boxes in these rules.

Table 2 shows that class G has more rules/boxes with smaller coverage (four rules that cover from one to seven cases with total 16 cases covered by these rules). These boxes can be called "*mini*" boxes. In contrast, class R (red) has no rules and boxes with such small coverage. It means a better generalization for R class, than for G class, in these rules. When rules are analyzed with large coverage, the situation is the opposite. The first three G rules cover 428 cases (96.4% of G cases), while the first two R rules (rules $R_2$ and $R_4$) cover 192 cases, 80.3% of R cases).

Next, the rules that cover a small number of cases are more complex. This is rather a memorization of cases, than their generalization. Such complex rules are needed only for a small number of cases. This is an indication that the domain experts, who designed attributes for WBC data correctly captured/engineered critical attributes. It also can indicate superior human abilities to generate informative features manually. While, deep learning algorithms can automatically discover informative features, often it is challenging to interpret and explain them.

The 10-fold cross validation of a comparable ID3 Decision Tree (DT) gave the average accuracy of 92.85 %, with min of accuracy equal to 89.86% and max of accuracy equal to 94.58% [9]. The representative DT of these DTs has 30 terminal nodes that contain seven or less cases [9]. In contrast the BC algorithm produced 13 boxes/rules, and *only four* of them contain seven or less cases on all WBC data.

Figs. 10-16 illustrate all boxes showing the cases, which cross these boxes. In addition, Figs. 12-16 show also the cases of other colors, which are *removed* before discovering a given box by requiring not to belong to a set of prior boxes, listed in the rules in Table 2. All these visualizations are done in accordance with the partial dynamic ILBC visualization method illustrated in Fig. 6a.

Example: in Fig. 10, 382 green cases are classified with box $B_1$. The BC algorithm removed theses 382 green cases before the next search. Because BC algorithm is sequential hierarchical process, it searches for each box in the sequential order such as $B_1$, then $B_2$ and then $B_3$ and so on. This means that after cases of box $B_1$ are removed, there will be only $444 - 382 = 62$ green cases remaining for the next search. Because of the small number of cases left, there is a good chance that a rule for them can be overfitting. Therefore, it will be beneficial to analyze all green cases without removing cases of the prior boxes to get another rule with higher coverage.

Figs. 10-11 represent situations when boxes are discovered *without preconditions* on other boxes. These pictures show boxes $B_1$, $B_2$, $B_3$, and $B_4$ with only cases that cross them. Figs. 12-16 represent situations with boxes discovered *with a precondition* that cases that cross prior boxes.

In these figures for each box $B_i$, the first picture shows only cases which cross box $B_i$ after removing cases from $B_1$-$B_4$ and the second picture also shows cases from the opposite class that cross $B_i$ without removing cases that cross prior boxes. For example, Fig. 12 first shows 14 red cases that cross box $B_5$ and then it shows both red and green cases without removing 382 cases that cross box $B_1$ and 28 cases that cross box $B_3$. For box $B_6$ the first picture shows 18 green cases, and the second picture shows both green and red cases without removing red cases from boxes $B_2$ and $B_4$.

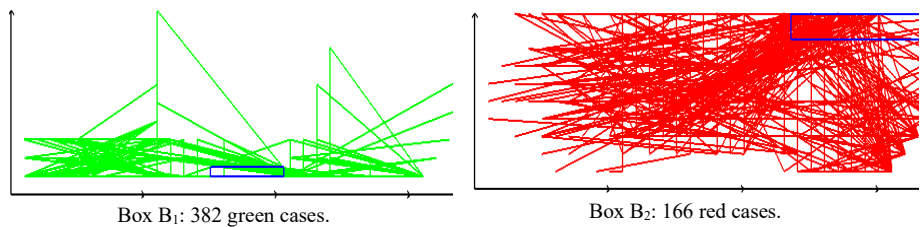

Box $B_1$: 382 green cases.          Box $B_2$: 166 red cases.

Fig. 10. Boxes $B_1$ and $B_2$.

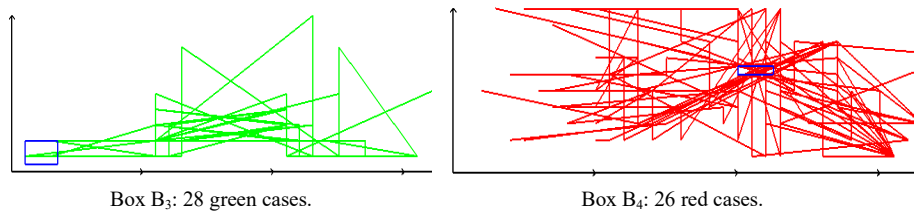

Box $B_3$: 28 green cases.          Box $B_4$: 26 red cases.

Fig. 11. Boxes $B_3$ and $B_4$.

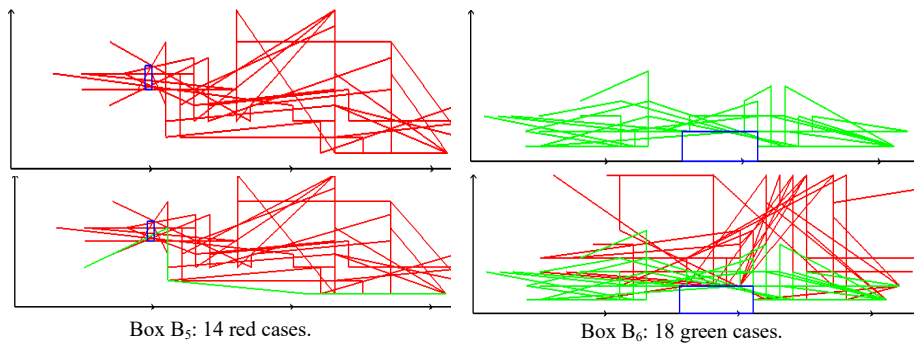

Box $B_5$: 14 red cases.          Box $B_6$: 18 green cases.

Fig. 12. Boxes $B_5$ and $B_6$.

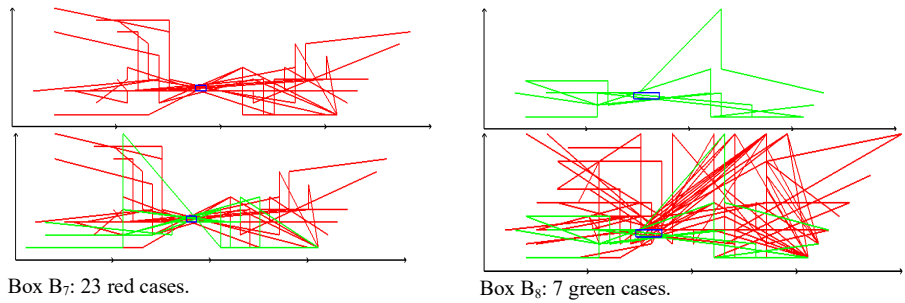

Box $B_7$: 23 red cases.　　　　　Box $B_8$: 7 green cases.
Fig. 13. Boxes $B_7$ and $B_8$.

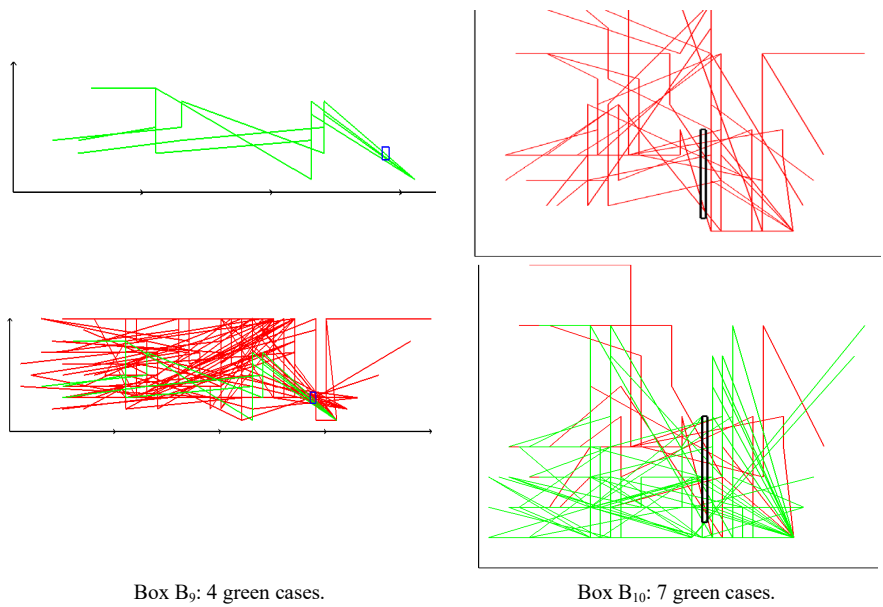

Box $B_9$: 4 green cases.　　　　　Box $B_{10}$: 7 green cases.
Fig. 14. Boxes $B_9$ and $B_{10}$.

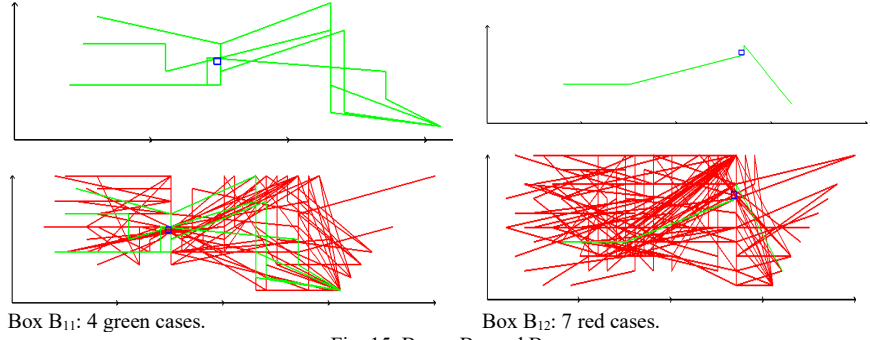

Box $B_{11}$: 4 green cases.   Box $B_{12}$: 7 red cases.

Fig. 15. Boxes $B_{11}$ and $B_{12}$.

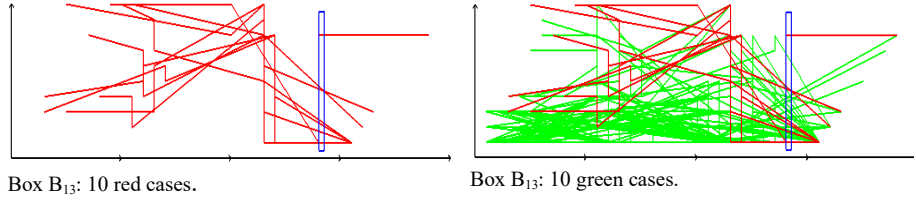

Box $B_{13}$: 10 red cases.   Box $B_{13}$: 10 green cases.

Fig. 16. Box $B_{13}$.

**Pruning**. Boxes $B_9$, $B_{11}$ with four green cases each and box $B_{12}$ with one green case can be used to prune the set of rules by creating modified rules,

$R_{9M}$: $\mathbf{x} \in B_9 \Rightarrow \mathbf{x} \in R$ (47 red /4 green).
$R_{11M}$: $\mathbf{x} \in B_{11} \Rightarrow \mathbf{x} \in R$ (28 red /4 green).
$R_{12M}$: $\mathbf{x} \in B_{12} \Rightarrow \mathbf{x} \in R$ (52 red /1 green).

Here $\mathbf{x} \in B_i$ means that polyline for n-D point $\mathbf{x}$ crosses the box $B_i$.

As it can be seen these rules have low error rate. Currently, the pruning process is interactive, therefore the end-users can explore them and accept if the error rate is tolerable.

**Joining rules.** The next task is decreasing the number of rules, which is demonstrated on 13 rules shown above. The proposed approach joins rules by combining them including the use of the else condition. In contrast with the pruning, this process does not introduce any error. The result is shown in Table 3.

The steps of the **Rule Joining (RJ) algorithm** are:

Step 1: Combine rules with a single rectangle of a given class.

Example: $R_{1,3}$: $\mathbf{x} \in B_1 \cup B_3 \Rightarrow \mathbf{x} \in G$ (410 cases).

Step 2: Find the opposite class rules that are conditioned by rectangles used in Step 1.

Example: Rule 5 is created after BC algorithm has removed 602 samples (410 green cases and 192 red cases) from boxes $B_1$, $B_2$, $B_3$, and $B_4$. Without removing these 602 samples, some green samples are crossing through box $B_5$. Therefore, rule $R_5$ is conditioned by $\mathbf{x} \notin B_1 \cup B_3$.

Step 3: Combine rules from Steps 1 and 2.

Example: $R_{1,3,5}$: $\mathbf{x} \in B_1 \cup B_3 \Rightarrow \mathbf{x} \in G$ (else $\mathbf{x} \in B_5 \Rightarrow \mathbf{x} \in R$) (428 cases). Here rule $R_5$ covers only 14 cases that can be viewed as a potential overfitting, while rule $R_{1,3,5}$ covers 428 cases. The else condition makes $R_5$ a part of the larger rule.

The analysis of rules in Table 3 shows that seven rules $R_{1,3,5}$, $R_{8,9}$, $R_{11,12}$, $R_{2,4,6,8}$, $R_7$, $R_{10}$, and $R_{13}$ are equivalent to 13 original rules. Here rules $R_{8,9}$, $R_{11,12}$, $R_{10}$, and $R_{13}$ cover 10 or less cases with total 36 cases (16 green and 20 red). Excluding these rules and, respectively refusing to classify cases that satisfy them will eliminate potential overfitting.

Table 3. Rules after joining.

| *Expanded benign (green, G) class rules.* |
|---|
| $R_{1,3}$: $\mathbf{x} \in B_1 \cup B_3 \Rightarrow \mathbf{x} \in G$ (410 cases) |
| $R_{1,3,5}$: $\mathbf{x} \in B_1 \cup B_3 \Rightarrow \mathbf{x} \in G$ (else $\mathbf{x} \in B_5 \Rightarrow \mathbf{x} \in R$) (424 cases) |
| $R_{8,9}$: $\mathbf{x} \in B_8 \cup B_9$ & $\mathbf{x} \notin B_2 \cup B_4 \cup B_5 \cup B_7 \Rightarrow \mathbf{x} \in G$ (11 cases) |
| $R_{11,12}$: $\mathbf{x} \in B_{11} \cup B_{12}$ & $\mathbf{x} \notin B_2 \cup B_4 \cup B_5 \Rightarrow \mathbf{x} \in G$ (5 cases) |
| *Expanded malignant (red, R) class rules.* |
| $R_{2,4}$: $\mathbf{x} \in B_2 \cup B_4 \Rightarrow \mathbf{x} \in R$ (192 cases) |
| $R_{2,4,6}$: $\mathbf{x} \in B_2 \cup B_4 \Rightarrow \mathbf{x} \in R$ (else $\mathbf{x} \in B_6 \Rightarrow \mathbf{x} \in G$) (210 cases) |
| $R_7$: $\mathbf{x} \in B_7$ & $\mathbf{x} \notin B_3 \cup B_6 \Rightarrow \mathbf{x} \in R$ (13 cases) |
| $R_{2,4,8}$: $\mathbf{x} \in B_2 \cup B_4 \Rightarrow \mathbf{x} \in R$ (else $\mathbf{x} \in B_8$ & $\mathbf{x} \notin B_5 \cup B_7 \Rightarrow \mathbf{x} \in G$) (199 cases) |
| $R_{2,4,6,8}$: $\mathbf{x} \in B_2 \cup B_4 \Rightarrow \mathbf{x} \in R$ (else $\mathbf{x} \in B_6 \cup B_8$ & $\mathbf{x} \notin B_5 \cup B_7 \Rightarrow \mathbf{x} \in G$) (217 cases) |
| $R_{10}$: $\mathbf{x} \in B_{10}$ & $\mathbf{x} \notin B_3 \cup B_6 \cup B_8 \cup B_9 \Rightarrow \mathbf{x} \in R$ (10 cases) |
| $R_{13}$: $\mathbf{x} \in B_{13}$ & $\mathbf{x} \notin B_1 \cup B_3 \cup B_6 \cup B_8 \cup B_9 \cup B_{11} \cup B_{12} \Rightarrow \mathbf{x} \in R$ (10 cases) |

## 4.2. Experiment with stratified 10-fold cross validation

All discovered WBC rules were presented above in Tables 2-3. This section analyzes rules $R_{1,3}$ and $R_{2,4}$ discovered by BC algorithm, which classify over 88% of WBC data. To get the even number of coordinates for the algorithm to run, the attribute $X_9$ was used twice. Table 4 presents results of BC algorithm on WBC data for these rules for *all* 444 green (benign) and 239 red (malignant) cases of WBC data without splitting data to training and validation subsets. The results are summarized in tables. Here, rule $R_{1,3}$ predicts green class and $R_{2,4}$ predicts red class based on all WBC data. Hyper-parameters of the rectangles $B_1$ and $B_3$ used in rule $R_{1,3}$ are $x_1=15$, $x_2=20.5$, $y_1=1$, $y_2=1.5$ for $B_1$ and $x_1=1$, $x_2=3.5$, $y_1=0.5$, $y_2=2$ for $B_3$. Hyper-parameters of the rectangles $B_1$ and $B_4$ used in rule $R_{2,4}$ are $x_1=23.5$, $x_2=39.5$, $y_1=8.5$, $y_2=10$ for $B_2$ and $x_1=20$, $x_2=22.5$, $y_1=6$, $y_2=6.5$ for $B_4$. These rules are:

$$R_{1,3}: \mathbf{x} \in B_1 \cup B_3 \Rightarrow \mathbf{x} \in G \text{ (Benign)}.$$

$$R_{2,4}: \mathbf{x} \in B_2 \cup B_4 \Rightarrow \mathbf{x} \in R \text{ (Malignant)}.$$

Table 4. Precision and recall of rules $R_{1,3}$ and $R_{2,4}$ for all data.

| Rule | WBC dataset | Rule precision (%) | Rule recall (%) |
|---|---|---|---|
| $R_{1,3}$ | 683 cases | 100% | 92.34% |
| $R_{2,4}$ | 683 cases | 100% | 80.33% |

Table 5. Precision and recall of rule $R_{1,3}$ and $R_{2,4}$ for stratified 10-fold cross validation (average and min max interval).

| Rule | Rule precision | | Rule recall | |
|---|---|---|---|---|
| | Training (%) | Validation (%) | Training (%) | Validation (%) |
| $R_{1,3}$ | 99.67 [99.2, 99,73] | 99.57 [99.65, 100] | 93.38 [93.25, 99,75] | 94.56 [90.91, 100] |
| $R_{2,4}$ | 95.90 [95.52, 96.5] | 96.80 [95.0, 100.0] | 89.21 [88.37, 89,76] | 85.00 [75.0, 100.0] |

The importance of the results in Table 4 is that two simple visual rules $R_{1,3}$, and $R_{2,4}$, which need only four unchanged rectangles $B_1$-$B_4$ allowed to recall of 80.3 and 92,34% cases with 100% precision on all data. Table 5 shows a similar performance in stratified 10-fold cross validation of BC algorithm on WBC data for rules $R_{1,3}$ ( green class) and rule $R_{2,4}$ (red class). For both rules WBC data were split to training and validation datasets in accordance with the total number of cases in the respective class.

## 5. Box Classification algorithm for multiclass unbalanced high-resolution data

The goal of designing a box classification algorithm for *multiclass imbalanced high-resolution data* is motivated by the Page Block Classification (PBC) dataset, which has all these properties. PBC dataset [6] has 5473 cases with 10 attributes each. There are 4913 cases from class Text (class $C_1$), 329 cases from class Horizontal Line (class $C_2$), 28 cases from class Graphic (class $C_3$), 88 cases from class Vertical Line (class $C_4$) and 115 cases from class Picture (class $C_5$). This dataset is *heavily imbalanced* in the number of cases of its *five classes* that range from 28 to 4913 cases. PBC is also a high-resolution dataset where each attribute has a large number of values, where some attributes have 5 digits in each value leading to $10^5$ values. Therefore, the BC algorithm is presented below in the terms of PBC data being applicable to many multiclass unbalanced high-resolution data. This expanded BC algorithm is divided into several sub-algorithms presented below.

**Divide and conquer (DC) algorithm for imbalanced data.** To classify imbalanced data, a common divide and conquer approach is used with three steps presented for five PBC data classes.

Step 1: *Combine* Classes $C_2$-$C_5$ into a single class $C_{2345}$. Classify cases between class $C_1$ and class $C_{2345}$. This task is less imbalanced with 4913 cases in $C_1$ and 560 cases in $C_{2345}$ than the task with 5 classes $C_1$-$C_5$.

Step 2: *Classify* cases between class $C_2$ and class $C_{345}$ that combines classes $C_3$-$C_5$ which is also less imbalanced: 329 cases in $C_2$ vs 231 cases in $C_{345}$.

Step 3: *Classify* 231 cases in joint class $C_{345}$ among three classes $C_3$, $C_4$ and $C_5$.

**DT Guided (DTG) algorithm for high-resolution dataset PBC data**. PBC is a *high-resolution dataset* with many digits on attributes. In contrast, in WBC data, each value consists of a single digit where the exhaustive grid search runs only on 1000 boxes. The exhaustive grid search in PBC data requires to run on a grid that is several orders of magnitude larger. So, the run time needs to be decreased. In [13] a random selection of grid cells (boxes) was used to decrease the search time. Here a *Decision Tree* (DT) is used as a **guide** for finding promising boxes. The steps of the DT Guided (DTG) algorithm are as follows.

Step 1: *Build a DT* on the same data.
Step 2: *Select high purity* DT branches, where a single class highly dominates.
Step 3: *Build boxes* based on those branches (one or more boxes from the branch).
Step 4: *Search for better boxes* in the vicinity of boxes derived from the DT step 3.

The link between branches of the DT and the boxes is shown in [9]. Each branch of DT can be a source of several boxes because boxes in static ILP are two-dimensional, while each DT branch commonly involve more attributes. The current implementation of step 3 starts from nodes of the branch that are close to the root of the DT because they typically cover more cases than the nodes, which are closer to the terminal nodes. As an illustration consider the following branch of the DT for the 4-D point $\mathbf{x}=(x_1,x_2,x_3,x_4)$ with all $x_i \in [0,10]$,

If $(x_1 > 5$ & $x_3 > 6)$ & $(x_2 < 3$ & $x_4 < 7)$ then $\mathbf{x} \in$ class 1.

Here box $B_1$ from $(x_1 > 5$ & $x_3 > 6)$ and box $B_2$ from $(x_2 < 3$ & $x_4 < 7)$ are created. Respectively $B_1$ is defined by $10 \geq x_1 > 5$, $10 \geq x_3 > 6$. When the limit 10 is a known it can be simplified to $(x_1 > 5$ & $x_3 > 6)$ as it is commonly written in the DTs.

**BC algorithm as a generalization of Decision Tree algorithm**. DT and rules based on boxes serve the same common goal of providing **interpretable** and easily **visualizable** models. The major limitation of models constructed by the DT algorithms is the need to select a start attribute (*tree root*). This *narrows the class of models* that can be discovered. It led to development of *Random Forests* (RFs) algorithms, where multiple DTs are combined by voting. RFs fundamentally expanded the class of models but with the cost of *losing interpretability*. The BC algorithm covers a *wider class of models* than DTs because they to do not require a root attribute. This is an *advantage* for the BC algorithm over the DTs.

**ILC box visualization as a richer visualization of Decision Tree**. The description of a DT as sequences of boxes is not only an alternative to *describe* DT but also an alternative to *visualize* it in ILC as boxes as Figs. 6-16 show. The advantage of this visualization of DT relative to a traditional visualization of decision trees is in ability to **trace each n-D point** in the tree visually. The traditional DT visualization is not doing this in contrast with the box visualization in ILC. It shows all cases that go through the boxes and fully represents the DT. It makes this visualization richer and more informative. ILC allows distinguishing DT branches by using distinct colors when all branches are visualized together or by showing each branch in the separate ILC.

## 6. Case study with Page Block Classification data in ILC2

### 6.1. Construction of the guiding decision tree

In accordance with the DT guided algorithm for PBC high-resolution data we (a) build a DT on the PBC data,

(b) select high purity DT branches, where a single class highly dominates,

(c) build boxes based on those branches, and

(d) search for better boxes in the vicinity of boxes from step 3.

Table 6 shows the characteristics of the constructed decision tree. To compute the *total precision of k rules* for a class we use the **weighted precision** formula [15]

$$P = \sum_{i=1}^{k}(p_i c_i)/\sum_{i=1}^{k} c_i$$

where $p_i$ is precision of rule $R_i$ and $c_i$ is the number of cases covered by $R_i$.

Example: Consider rule $R_a$ that predicted 100 cases with 90 correctly classified and 10 misclassified cases ($P_a$=90% precision). Another rule $R_b$ classified 200 cases, with 160 cases correctly and 40 incorrectly classified ($P_2$=80% precision).

Here $c_a$=100 and $c_b$=200, $W_a=c_a/(c_a+c_b)\approx 0.33$ and $W_b=c_b/(c_a+c_b)\approx 0.67$ and the weighted precision $P = 0.33 * 90\% + 0.67 * 80\% = 83.33\%$. Table 6 uses this concept of **weighted precision** $P$ that is the weighted sum of precisions of all rules $R_1$-$R_k$:

$$P = W_1 P_1 + W_2 P_2 + \cdots + W_k P_k.$$

where the weight $W_i$ of each rule $R_i$ is computed as follows:

$$W_i = \frac{c_i}{(c_1 + c_2 \ldots c_k)}.$$

where $c_i$ is the total number of cases classified by rule $R_i$ as above.

The abilities to use a decision tree as a good guidance heavily depend on the quality of the decision tree, therefore, we need to analyze this tree. An ideal DT will be highly accurate without overfitting. Much more often a DT is not perfect but with insufficient accuracy and significant overfitting.

Multiple DT branches may have terminal nodes, which include just few cases rather memorizing data than learning the patterns. Table 6 demonstrates highly insufficient precision of the constructed DT for some data subsets. *None* of the cases from class 3 is correctly classified.

This means that the guidance from the decision tree can be incomplete. The hope is that the DT guidance will decrease the number of cases and the areas to search for good boxes needs dramatically. It will allow an exhaustive search with the remaining parts of the visualization data space.

Table 6. Weighted precision for all classes with DT for PBC dataset.

| Class | Weight (%) | Precision (%) | Classified cases | Weighted precision (%) |
|---|---|---|---|---|
| | **Training** | | | |
| Class 1 | 89.14% | 98.05% | 3950 | 87.41% |
| Class 2 | 7.15% | 74.76% | 317 | 5.35% |
| Class 3 | 0.00% | 0.00% | 0 | 0.00% |
| Class 4 | 1.62% | 90.28% | 72 | 1.47% |
| Class 5 | 2.08% | 46.74% | 92 | 0.97% |
| All | 100.00% | | 4431 | 95.19% |
| | **Validation** | | | |
| Class 1 | 58.10% | 97.56% | 287 | 56.68% |
| Class 2 | 7.89% | 58.97% | 39 | 4.66% |
| **Class 3** | **0.00%** | **0.00%** | **0** | **0.00%** |
| Class 4 | 31.17% | 5.19% | 154 | 1.62% |
| Class 5 | 2.83% | 21.43% | 14 | 0.61% |
| All | 100.00% | | 494 | 63.56% |
| | **Testing** | | | |
| Class 1 | 87.77% | 98.75% | 481 | 86.68% |
| Class 2 | 7.85% | 72.09% | 43 | 5.66% |
| Class 3 | 0.00% | 0.00% | 0 | 0.00% |
| Class 4 | 1.64% | 77.78% | 9 | 1.28% |
| Class 5 | 2.74% | 60.00% | 15 | 1.64% |
| All | 100.00% | | 548 | 95.26% |

Our experiments with PBC dataset reported in the next section shows success of this strategy. An alternative approach is building a better tree or pruning the available tree. While pruning is a common way to decrease overfitting of DT it can decrease accuracy.

In the next section we show that BC algorithm allowed getting high accuracy without pruning because it is not limited by the tree structure.

## 6.2. Construction rules in stratified 10-fold cross validation experiment

For stratified 10-fold cross validation, PBC dataset is split into 90%:10% where 90% used for into training and validation set and 10% for independent testing. Training and validation set was then split to 90% training set and 10% validation set.

Tables 7 to 11 present the result of BC algorithm *1$^{st}$ fold* in stratified 10-fold cross validation. Results for folds 2-10 are summarized in Table 12. Table 7 presents hyper-parameters of rectangles *derived from the DT* for this fold and table 8 presents rules based on these rectangles. This DT was build using respective 90% of training and validation set designated for this fold.

Table 7. Hyper-parameters of the rectangles $B_1$-$B_{11}$ (BC algorithm, PBC dataset) of 1st fold in stratified 10-fold cross validation.

| Box | Hyper-parameters | Box | Hyper-parameters |
|---|---|---|---|
| $B_1$ | $X_6 < 0.0011$ & $0.0015 \leq X_0 < 0.0214$ | $B_2$ | $0.1550 \leq X_4 < 0.9394$ & $0.0065 < X_0 \leq 0.107$ |
| $B_3$ | $0.7525 \leq X_5 < 1$ | $B_4$ | $0.0750 \leq X3 < 1$ & $0.0001 \leq X6 < 1$ |
| $B_5$ | $0 \leq X3 < 0.0005$ | $B_6$ | $0.3730 \leq X_4 < 1$ & $0.0115 \leq X_3 < 0.537$ |
| $B_7$ | $0 \leq X_4 < 0.12$ | $B_8$ | $0 < X_3 \leq 0.0005$ |
| $B_9$ | $0 < X_3 \leq 0.0005$ | $B_{10}$ | $0 \leq X_4 \leq 0.2944$ |
| $B_{11}$ | $0.006 \leq X_2 \leq 0.6058$ | | |

Table 8. Rules $R_1$-$R_6$ (BC algorithm, PBC dataset) using boxes $B_1$-$B_{11}$.

| | |
|---|---|
| *Class $C_1$ rule* | $R_1$: $\mathbf{x} \in B_1 \cup B_2 \cup B_3 \Rightarrow \mathbf{x} \in$ Class $C_1$ |
| *Class $C_{2,3,4,5}$ rule* | $R_2$: $\mathbf{x} \in B_4 \cup B_5 \cup B_6 \cup B_7$ & $\mathbf{x} \notin B_1 \cup B_2 \cup B_3 \Rightarrow \mathbf{x} \in$ Class $C_{2,3,4,5}$ |
| *Class $C_2$ rule* | $R_3$: $\mathbf{x} \in B_8 \Rightarrow \mathbf{x} \in$ Class $C_2$ |
| *Class $C_4$ rule* | $R_4$: $\mathbf{x} \in B_9$ & $\mathbf{x} \notin B_8 \Rightarrow \mathbf{x} \in$ Class $C_4$ |
| *Class $C_5$ rule* | $R_5$: $\mathbf{x} \in B_{10}$ & $\mathbf{x} \notin B_8 \cup B_9 \Rightarrow \mathbf{x} \in$ Class $C_5$ |
| *Class $C_3$ rule* | $R_6$: $\mathbf{x} \in B_{11}$ & $\mathbf{x} \notin B_8 \cup B_9 \cup B_{10} \Rightarrow \mathbf{x} \in$ Class $C_3$ |

The process of classification using these rules is hierarchical Case **x** is classified between class $C_1$ and a joined class $C_{2,3,4,5}$ and then if **x** is in $C_{2,3,4,5}$ then **x** is classified to classes $C_2$-$C_5$. See Fig. 17.

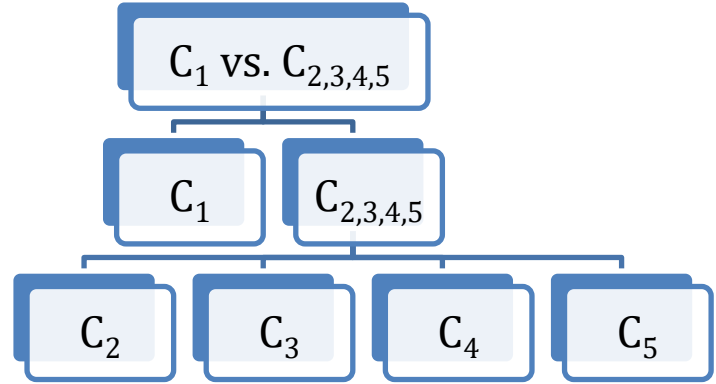

Fig. 17. Classification process.

Table 9 shows the number of cases that were predicted for training, validation, and testing sets in table 8. The notation $C_i \Rightarrow C_i$ indicates situations when cases of class $C_i$ were predicted correctly as cases of class $C_i$, e.g., $C_{2345} \Rightarrow C_{2345}$ means that cases of the joint class $C_{2345}$ were predicted correctly as cases of this joint class. The notation $C_i \Rightarrow C_j$ when $i \neq j$ indicates situations when cases of class $C_i$ were predicted *incorrectly* as cases of class $C_j$.

Table 10 shows precision and recall percentages of all rules $R_1$-$R_6$ on training, validation, and testing sets.

Table 9. Number of cases that satisfy rules $R_1$-$R_6$ (BC algorithm, PBC dataset) of 1st fold in stratified 10-fold cross validation.

| Rule | Training | | Validation | | Testing | |
|---|---|---|---|---|---|---|
| $R_1$ | $C_1 \Rightarrow C_1$ | $C_{2345} \Rightarrow C_1$ | $C_1 \Rightarrow C_1$ | $C_{2345} \Rightarrow C_1$ | $C_1 \Rightarrow C_1$ | $C_{2345} \Rightarrow C_1$ |
|  | 3782 | 55 | 420 | 5 | 445 | 5 |
| $R_2$ | $C_{2345} \Rightarrow C_{2345}$ | $C_1 \Rightarrow C_{2345}$ | $C_{2345} \Rightarrow C_{2345}$ | $C_1 \Rightarrow C_{2345}$ | $C_{2345} \Rightarrow C_{2345}$ | $C_1 \Rightarrow C_{2345}$ |
|  | 359 | 35 | 42 | 4 | 48 | 9 |
| $R_3$ | $C_2 \Rightarrow C_2$ | $C_{345} \Rightarrow C_2$ | $C_2 \Rightarrow C_2$ | $C_{345} \Rightarrow C_2$ | $C_2 \Rightarrow C_2$ | $C_{345} \Rightarrow C_2$ |
|  | 251 | 2 | 28 | 0 | 32 | 3 |
| $R_4$ | $C_4 \Rightarrow C_4$ | $C_{235} \Rightarrow C_4$ | $C_4 \Rightarrow C_4$ | $C_{235} \Rightarrow C_4$ | $C_4 \Rightarrow C_4$ | $C_{235} \Rightarrow C_4$ |
|  | 66 | 6 | 7 | 0 | 7 | 1 |
| $R_5$ | $C_5 \Rightarrow C_5$ | $C_{234} \Rightarrow C_5$ | $C_5 \Rightarrow C_5$ | $C_{234} \Rightarrow C_5$ | $C_5 \Rightarrow C_5$ | $C_{234} \Rightarrow C_5$ |
|  | 83 | 8 | 11 | 1 | 12 | 0 |
| $R_6$ | $C_3 \Rightarrow C_3$ | $C_{245} \Rightarrow C_3$ | $C_3 \Rightarrow C_3$ | $C_{245} \Rightarrow C_3$ | $C_3 \Rightarrow C_3$ | $C_{245} \Rightarrow C_3$ |
|  | 22 | 2 | 3 | 1 | 3 | 0 |

The rules $R_1$-$R_6$ do not cover the same number of cases, therefore the balancing of their contribution to the total precision of requires using the **weighted precision** that will account this difference. The formula for the weighted precision is given in the previous section.

The rule $R_2$ is excluded from computing the weighted precision because it is intermediate rule, which predicts a joined class $C_{2,3,4,5}$ is and only the rules, which predict actual classes $C_1$-$C_5$ called terminal level rules are included. The main steps to calculate weighted precision for PBC dataset are below.

Step 1: Calculate weighted precision of the terminal level rules $R_1$, $R_3$, $R_4$, $R_5$, and $R_6$ that predict actual classes $C_1$, $C_2$, $C_4$, $C_5$, and $C_3$, respectively.

Step 2: Calculate weighted precision of $R_1$, $R_3$, $R_4$, $R_5$, and $R_6$.

Step 3: Sum up the weighted precisions from Step 2.

The results for 1st fold in stratified 10-fold cross validation are presented in Table 11, where the weighted precision of all rules is 98.29% for the training set, 98.53% for the validation set, and 98.23% for the testing set. The summary of other folds is in Table 12. It shows that average precision of BC algorithm 10-fold cross validation for PBC dataset is 97.30%.

Table 11. Weighted precision for all classes of BC algorithm for PBC dataset of 1$^{st}$ fold in stratified 10-fold cross validation.

| Rule | Weight (%) | Precision (%) | Classified cases | Weighted precision (%) |
|---|---|---|---|---|
| | **Training** | | | |
| $R_1$ | 89.71% | 98.57% | 3837 | 88.43% |
| $R_3$ | 5.92% | 99.21% | 253 | 5.87% |
| $R_4$ | 1.68% | 91.67% | 72 | 1.54% |
| $R_5$ | 2.13% | 91.21% | 91 | 1.94% |
| $R_6$ | 0.56% | 91.67% | 24 | 0.51% |
| All | 100.00% | | 4277 | 98.29% |
| | **Validation** | | | |
| $R_1$ | 89.29% | 98.82% | 425 | 88.24% |
| $R_3$ | 5.88% | 100.00% | 28 | 5.88% |
| $R_4$ | 1.47% | 100.00% | 7 | 1.47% |
| $R_5$ | 2.52% | 91.67% | 12 | 2.31% |
| $R_6$ | 0.84% | 75.00% | 4 | 0.63% |
| All | 100.00% | | 476 | 98.53% |
| | **Testing** | | | |
| $R_1$ | 88.58% | 98.89% | 450 | 87.60% |
| $R_3$ | 6.89% | 91.43% | 35 | 6.30% |
| $R_4$ | 1.57% | 87.50% | 8 | 1.38% |
| $R_5$ | 2.36% | 100.00% | 12 | 2.36% |
| $R_6$ | 0.59% | 100.00% | 3 | 0.59% |
| All | 100.00% | | 508 | 98.23% |

Table 12. Precision for stratified 10-fold cross validation of BC algorithm for PBC dataset.

| Fold's number | Weighted precision (%) for testing data |
|---|---|
| 1$^{st}$ fold | 98.23% |
| 2$^{nd}$ fold | 96.65% |
| 3$^{rd}$ fold | 98.62% |
| 4$^{th}$ fold | 97.16% |
| 5$^{th}$ fold | 97.84% |
| 6$^{th}$ fold | 97.64% |
| 7$^{th}$ fold | 96.36% |
| 8$^{th}$ fold | 96.18% |
| 9$^{th}$ fold | 97.16% |
| 10$^{th}$ fold | 97.11% |
| Average | 97.30% |

## 6.3. Comparison of Box Classification with Decision Tree and published results

The exact comparison of our results with published results is difficult because the authors use different splits of data. Table 13 presents results of comparisons with clearly stated what is the difference in conducted experiments. These differences are discussed in detail below.

The KNN result is obtained with 80:20 split. With the KNN method, the precision is 93.51%, which is lower than our BC algorithm at 97.30%. However, this lower precision is to be expected because typically precision with 90:10 split is greater than with 80:20 split. The C4.5 DT precision is obtained with 90:10 split in 10-fold cross validation.

Therefore, C4.5 DT result would give us a better comparison with BC algorithm because PBC dataset is split in the same ratio. With BC algorithm precision is slightly higher at 97.30% compared to C4.5 DT precision at 96.95%. Table 13 also presents C4.5 DT precision with all 100% of PBC data used as training data, ID3 DT with a single 81%:9%:10% split of PBC data, and Block Classification with 10-fold cross validation 81%:9%:10% of PBC data.

Both C4.5 DT and ID3 DT did not correctly classify any cases from class 3, indicating insufficient accuracy and precision. The BC algorithm **classified all classes with higher precision on independent test data**. Furthermore, BC algorithm with In-line coordinate can visualization of PBC dataset allowing the end-user easier and faster understanding data.

Table 13. Comparisons with published results for PBC dataset

| Algorithm | Precision (%) on training data | Precision (%) on validation data | Precision (%) on test data | Classes with 0% precision (completely misclassified) |
|---|---|---|---|---|
| K-nearest Neighbor with a single 80%:20% training-validation split [23] | Not reported | 93.51 | Precision % on validati-on data | Not reported |
| C4.5 Decision Tree with 10-fold cross validation 90%:10% training-validation split [24] | Not reported | 96.95** | Not reported | Not reported |
| C4.5 Decision Tree with 100% training data | 96.02% | N/A | N/A | Class 3 |
| ID3 Decision Tree with a single 81%:9%:10% split | 95.19% | 63.56% | 95.26% | Class 3 |
| **Block Classification** with 10-fold cross validation 81%:9%:10% | **98.26%** | **96.34%*** | **97.30%*** | No such classes |

*average; **presumed average.

Table 14 is produced as follows. Out of all data, 90% were selected randomly for training and validation and 10% for testing. Then the first 90% of data was split in 90:10 ratio for training and validation data. The precision of all classes in training is 95.19% and 95.26% in testing. Table 14 shows that precision of DT for PBC dataset is 95.26% which is slightly lower BC algorithm. In table 14, it can also be seen that the disadvantage of using DT for PBC dataset such that DT cannot classify any cases from class 3. With BC algorithm for PBC dataset, it was classified that all 5 classes with precision higher than 90% compared to DT algorithm where class 2,3,4, and 5 have precision lower than 80%. BC algorithm with In-line Coordinates also allows us to show how high dimensional dataset like PBC can be visualized on Cartesian coordinates compared to non-visualization method of DT.

Table 14. Weighted precision for all classes with Decision Tree ID3 for PBC dataset for a single 81%:9%:10% training: validation: test split.

| Class | Weight (%) | Precision (%) | Classified cases | Weighted precision (%) |
|---|---|---|---|---|
| | | *Training* | | |
| Class 1 | 89.14% | 98.05% | 3950 | 87.41% |
| Class 2 | 7.15% | 74.76% | 317 | 5.35% |
| Class 3 | **0.00%** | **0.00%** | **0** | 0.00% |
| Class 4 | 1.62% | 90.28% | 72 | 1.47% |
| Class 5 | 2.08% | **46.74%** | 92 | 0.97% |
| All | 100.00% | | 4431 | 95.19% |
| | | *Validation* | | |
| Class 1 | 58.10% | 97.56% | 287 | 56.68% |
| Class 2 | 7.89% | 58.97% | 39 | 4.66% |
| Class 3 | **0.00%** | **0.00%** | 0 | 0.00% |
| Class 4 | 31.17% | 5.19% | 154 | 1.62% |
| Class 5 | 2.83% | 21.43% | 14 | 0.61% |
| All | 100.00% | | 494 | 63.56% |
| | | *Testing* | | |
| Class 1 | 87.77% | 98.75% | 481 | 86.68% |
| Class 2 | 7.85% | 72.09% | 43 | 5.66% |
| Class 3 | **0.00%** | **0.00%** | 0 | 0.00% |
| Class 4 | 1.64% | 77.78% | 9 | 1.28% |
| Class 5 | 2.74% | 60.00% | 15 | 1.64% |
| All | 100.00% | | 548 | 95.26% |

## 7. Interpretability

A major advantage of the Box algorithm on the static ILC visualization is in **direct interpretability** of boxes and rules it produces when the input attributes are interpretable themselves. The limitation of this Box algorithm is in capturing only relatively simple interval rules.

For data with more complex patterns, we proposed the dynamic ILC Box algorithms, where boxes depend on the value of several attributes. Boxes at the very beginning of the 2-D visualization space capture dependences of just a few attributes but the boxes at the end capture dependences of all attributes. It allows capturing a variety of simple and complex relations. These relations are varied depending on the sequences and locations of attributes in the visualization space. It leads to combinatorial exploration of different sequences and locations of attributes in the visualization space to capture complex relations that are specific for a given data. Therefore, these dynamic algorithms are flexible but limited computationally.

Each box found by ILC-bases algorithms can be identified by the values of two functions: $f_{hor}$ and $f_{ver}$, which identify its low left corner. For the fully dynamic ILC in Fig. 5c, the horizontal function $f_{hor} = a+b+c$, is the sum of attributes/coordinates, which are located horizontally, and the vertical function $f_{ver}=d+e+f+g$ is the sum of values of the attributes/coordinates located vertically. In general, for other dynamic

ILC-based algorithms the exact mathematical class of rules is a subject of further exploration.

To justify that the boxes identified by linear functions are **interpretable** we use the method presented in [25] in this volume. The goal is to explain why a new n-D point **c** in the box B of a given class L belongs to that class L. We do this by finding two n-D points **a** and **b** in this box B, such that **c** is between them, **a**≤ **c** ≤**b**, i.e., for each $i=1{:}n$  $a_i \leq c_i \leq b_i$. The points **a** and **b** can be actual training points, or artificial ones. If actual such points exist then we output them in addition to artificially created points that can cover wider data intervals. All these n-D points are visualized in ILC-based space along with the box B to provide easy to capture visual explanation. When the rule contains several boxes the same is done for all these boxes.

A common **external method** today to explain a **non-interpretable model** $F(\mathbf{x})$ is using an explainable model $G(\mathbf{x})$ like a decision three to approximate $F(\mathbf{x})$ **locally** around n-D point **x**. We expand the pool of such external explainable models $G(\mathbf{x})$ with rules generated by the Box algorithm in the ILC-based lossless visualization spaces.

The approximation process with the Box algorithm is as follows. Consider a new n-D point **c** and prediction of its class by computing $F(\mathbf{c})$. This prediction by non-interpretable $F$ needs to be explained.

Step 1: Set up an initial grid resolution $r$ for the selected ILC-based lossless visualization space $S$.
Step 2: Plot all training data in $S$.
Step 3: Find all boxes $\{B\}$, which **c** belongs, i.e., boxes where the polyline **c*** of n-D point **c** crosses or ends.
Step 4: Find all $\{B\}$ with a dominant class purity greater or equal to the threshold $T$.
Step 5: If no such $B$ exist decrees the grid $r$ by value $\Delta$.
Step 6: Repeat steps 4-5 until a box with purity$(B) \geq T$ will be found. If no box found stop.
Step 7: If required box $B$ exists find two *training* n-D points **a** and **b** in $B$ from class L, such that **c** is between them, **a**≤ **c** ≤**b**, i.e., $\forall i=1{:}n$  $a_i \leq c_i \leq b_i$.
Step 8: If required box $B$ exists generate two *artificial* n-D points **d** and **e** in $B$, such that **c** is between them, **d**≤ **c** ≤**e**, i.e., $\forall i=1{:}n$  $d_i \leq c_i \leq e_i$.
Step 9. Visualize **a**,**b**,**c**, **d**,**e** in ILC-based space along with the box B.

If several such boxes exist then we will have multiple explanations. If none of such boxes exist then a more complex areas than simple boxes like defined by $B_1$&(not$B_2$) can be explored to be pure enough.

Can we get a more interesting or richer explanation with the Box algorithm than with a decision tree? A local DT explanation for **c** means tracing a DT branch that contains **c**. Each DT node conducts a simple test like $x_i \leq T_i$. The Box algorithm tests the box, i.e., two intervals like, $T_{1x} \leq x \leq T_{2x}$ and $T_{1y} \leq y \leq T_{2y}$ for x, y coordinates of the visualization plot grid. These two intervals are equivalent to a DT with a single branch with 4 nodes:

$$T_{1x} \leq x \ \& \ x \leq T_{2x} \ \& \ T_{1y} \leq y \ \& \ y \leq T_{2y}$$

For the static ILC $x$ and $y$ are original data attributes $x_i$. The rules, which include more than one box, like R$_{1,3}$: $\mathbf{x} \in$ B$_1 \cup$B$_3 \Rightarrow \mathbf{x} \in$ G from Table 3, will require several such small DTs with a single branch each. These DTs can be completely independent in contrast with branches of the DT which have a common root. It means that they can capture properties that an individual decision tree **cannot capture**. In the decision tree theory this limitation was lifted by building a set of decision trees in random forests algorithms.

Another fundamental distinct property of boxes is that the polyline (graph) $\mathbf{x}^*$ of the n-D point $\mathbf{x}$ can go through that box, not having any node of $\mathbf{x}^*$ inside of the box B. The rules based on such boxes express the properties **two nodes** of $\mathbf{x}^*$ that form a segment of the polyline that go through the box B. Each of this node require more than one attribute. This means that this box B captures relations between **two or more attributes**. Such properties are not captured by decision trees at least directly.

The **dynamic** ILC-based visualizations allow to express more complex models than static ILCs, because here coordinates of the visualization space x and y are functions of original attributes $x_1, x_2, \ldots, x_n$ not these attributes themselves.

In contrast with the random forest (RF) models the rules produced by the Box algorithm do not require voting, which create interpretability problem for RF. Thus, rules produced by the Box algorithm have a form of **general logical rules** on intervals of values without requiring a root attribute as it is for each DT. The abilities to produce all general logical rules on intervals is a subject of the further research. Major current **limitations** of the Box algorithm are computational to deal with the data of large dimensions, high resolution, and a large number of cases.

In summary the box algorithm can be used like a decision tree: (a) to **build interpretable models from scratch** and (b) to **explain locally any black box model** with advantages that it can capture **more general** and complex relations than Decision trees.

## 8. Conclusions

In this research, a novel 2-dimensional machine learning (2-D ML) methodology in the context of inline coordinates was developed. This is a full machine learning approach, which does not necessitate dealing with n-dimensional data in n-dimensional abstract space. It functions without any information loss with n-dimensional data in a 2-dimensional visualization space. Several static and dynamic alternative in-line-based coordinates have been developed for this. Following that, techniques for classification and regression based on these inline coordinates were introduced. A successful case study using WBC data showed that the strategy is feasible.

With a new methodology, this work advances full 2-D machine learning to become a new research area. It benefits from the ability to actively involve end users/domain exerts in the machine learning discovery process and model justification.

It enables the delivery of ML models that are comprehensible.

With In-line Coordinates data visualization, this study has shown the power of interpretable data classification techniques that are implemented in automatic and interactive modes, which have been tested with WBC and PBC datasets. The proposed BC algorithm allowed to successfully classify WBC and PBC datasets.

It was observed that BC algorithm worked well with lower feature resolution lesser for a higher resolution and larger dataset. Higher dimension of features makes finding the best order of coordinates difficult because this process require a great amount of time. This led to further development of BC algorithm with using a Decision Tree for guidance. It allowed finding an efficient order of coordinates and a set of boxes in a practical run time.

With In-line Based Coordinates, it was demonstrated that the power of visualization can reduce the overgeneralization of hyper-parameters produced by the guiding DT algorithm. The BC algorithm showed that it is possible to achieve better results compared to both published results of KNN and C4.5 DT algorithms. In comparison with the DT, the BC algorithm did not miss any class on highly imbalanced PBC dataset.

Further work will focus on improving algorithms for greater accuracy and expanding the range of potential patterns that can be found in in-line coordinates. Boxes can be substituted by a more complex interpretable structure. Multithreading with GPU will be developed to deal with larger datasets.